\documentclass[letterpaper, 16pt]{article}
\usepackage{booktabs}
\usepackage{color,soul}
\usepackage{amsbsy}
\usepackage{amsfonts}
\usepackage{caption}
\usepackage{tabularx} 
\usepackage{amsmath}  
\usepackage{graphicx} 
\usepackage[margin=1in,letterpaper]{geometry} 
\usepackage{cite} 
\usepackage[final]{hyperref} 
\hypersetup{
	colorlinks=true,       
	linkcolor=blue,        
	citecolor=blue,        
	filecolor=magenta,     
	urlcolor=blue         
}

\begin{document}
\title{\textbf{Progress Estimation and Phase Detection for Sequential Processes \footnote{Accepted by IMWUT/Ubicomp 2017}}} 
\author{Xinyu Li, Yanyi Zhang, Jianyu Zhang, Yueyang Chen, Shuhong Chen,\\ Yue Gu, Moliang Zhou and Ivan Marsic\\
{\tt\small\{xl264, yz593, jz549, yc788, sc1624, yg202, mz330, marsic\}@rutgers.edu}\\\\
Richard A. Farneth and Randall S. Burd\\
{\tt\small\{rfarneth, rburd\}@childrensnational.org}}

%
\maketitle

\begin{abstract}
Process modeling and understanding are fundamental for advanced human-computer interfaces and automation systems. Most recent research has focused on activity recognition, but little has been done on sensor-based detection of process progress. We introduce a real-time, sensor-based system for modeling, recognizing and estimating the progress of a work process. We implemented a multimodal deep learning structure to extract the relevant spatio-temporal features from multiple sensory inputs and used a novel deep regression structure for overall completeness estimation. Using process completeness estimation with a Gaussian mixture model, our system can predict the phase for sequential processes. The performance speed, calculated using completeness estimation, allows online estimation of the remaining time. To train our system, we introduced a novel rectified hyperbolic tangent (rtanh) activation function and conditional loss. Our system was tested on data obtained from the medical process (trauma resuscitation) and sports events (Olympic swimming competition). Our system outperformed the existing trauma-resuscitation phase detectors with a phase detection accuracy of over 86\%, an F1-score of 0.67, a completeness estimation error of under 12.6\%, and a remaining-time estimation error of less than 7.5 minutes. For the Olympic swimming dataset, our system achieved an accuracy of 88\%, an F1-score of 0.58, a completeness estimation error of 6.3\% and a remaining-time estimation error of 2.9 minutes.
\end{abstract}

\section{Introduction}
\label{sec:01}

A sequence of activities that achieves a certain goal represents a work process. For instance, the trauma resuscitation process consists of pre-arrival, patient-arrival, primary survey, secondary survey, post-secondary survey, and patient-leave phases; each phase consists of one or more activities. Several successful systems have been introduced for image classification \cite{simonyan2014very} and activity recognition \cite{li2016deep}. We take a further step in this analysis by developing a system for real-time process progress estimation. Online progress estimation has applications for many real-world problems, including automated human-computer interaction systems. For example, online detection of a sports process phase can be used to guide the camera on overhead drones for the video broadcasting. Online progress information for a medical process, such as the process completeness and remaining time, can help medical providers organize resources and schedule treatment timing. We first analyzed the trauma resuscitation process and the Olympic swimming process, each with six process phases (Table \ref{tab:01}). Then, we designed and evaluated our sensor-based system to estimate the work progress in three ways:
\begin{enumerate}
	\item \textbf{Process completeness,} indicating the percentage completion of the whole process.
	
	\item \textbf{Process phase,} indicating a major stage of the process progress. A phase may contain one or more logically-grouped activities that achieve a larger goal.
	
	\item \textbf{The remaining-time,} representing the estimated time left until the process finishes.
\end{enumerate}

\begin{table}[!htb]
	\centering
	\caption{The phase definitions for the trauma for trauma resuscitation dataset (left) and Olympic swimming dataset (right)}
	\captionsetup{justification=centering}
	\label{tab:01}
	\setlength{\belowrulesep}{0pt}
	\setlength{\aboverulesep}{0pt}
	\begin{minipage}{\columnwidth}
		\begin{center}
			\begin{tabular}{|p{0.18\textwidth}|p{0.3\textwidth}||p{0.20\textwidth}|p{0.23\textwidth}|}
				\toprule
				\textbf{Trauma Resuscitation Phases}                            & \textbf{Starts When}                             & \textbf{Swimming Competition Phases }                            & \textbf{Definition}            \\ \midrule
				Pre-Arrival (PA)                                & First trauma team member enters room      & Pre-Competition (PC)                        & Statement of participants                  \\ \midrule
				Patient Arrival (A)                             & Patient first appears in camera view      & Introduction (I)                            & Athletes enter the field with introduction \\ \midrule
				Primary (P)                                     & First ``primary survey activity'' starts   & Preparation (P)                             & Athletes prepare for swimming              \\ \midrule
				Secondary (S)                                   & First ``secondary survey activity'' starts & Competition (C)                             & Athletes compete                           \\ \midrule
				Post-Secondary (PS)                             & First ``post-secondary survey activity'' starts   & Replay (RP)                                 & Replay of the winner                       \\ \midrule
				Patient Leave (PL)                              & Patient leaves the room                   & Result (RE)                                  & Statement of results                       \\ \bottomrule
			\end{tabular}
		\end{center}
	\end{minipage}
\end{table}

Work processes can be roughly categorized into two types: sequential and parallel. Our current work does not model parallel processes. \textit{Linear} sequential processes can be partitioned into a set of ``phases" that occur one after another in a fixed order. For example, during the trauma resuscitation process or the sports event broadcasting process, the phases are rarely skipped or duplicated. \textit{Nonlinear} sequential processes allow phases to be repeated or performed in an arbitrary order. For example, when making a salad, the mixing, vegetable chopping, and sauce preparation can be performed in any order before serving. We focused on characterizing linear processes, although we demonstrated (Section \ref{sec:05}) that our system could be modified to handle nonlinear processes.

Previous process-phase detectors can be roughly categorized into three types: manual, shallow-modeled, and deep-learning-based. The first type used manually-generated event logs or medical equipment signals \cite{forestier2015automatic,padoy2012statistical}. As these systems worked with relatively noise-free datasets, they achieved good performance. However, manual detectors require manual log generation or signals from specific medical equipment, making them hard to implement and generalize. The use of sensors addressed these two issues at a cost: sensor data may be hard to obtain (e.g., wearable sensors are not preferred in medical settings as they may interfere with the work), and are subject to the hardware or environmental noise. The second type of phase detectors used shallow classifiers to predict phases from noisy sensor data \cite{forestier2015automatic,blum2008modeling,primus2016temporal}. Yet, the features and model selection were often arbitrary and difficult to generalize, because different researchers using different features can only claim that those features worked best for their specific scenarios \cite{li2016deep}. Due to its success in the image classification and speech recognition, deep learning and automatic feature extraction has given rise to the third type of phase detectors. All three approaches treated the process phase detection as a discrete classification problem, which overlooks the continuous nature of the processes and misses the associations between process phases and percentage completion. These limitations have caused systems to return logically impossible predictions \cite{li2016online}.

We address these issues and estimate process completeness as a continuous variable. Our system is designed to work with sensor data, as opposed to manually-generated event logs. We designed a multimodal structure to make the system compatible with different types of input sources. Features in our model are automatically learned using a deep neural network. We introduce a deep regression model to continuously estimate process completeness, and the rectified hyperbolic tangent (\textit{rtanh}) activation function to bound the regression output values in the context of process completeness. The system detects the current process phase based on the currently estimated process completeness using a Gaussian mixture model (GMM). Our model for completeness estimation is trained on the regression error as well as a novel conditional loss from the phase prediction. Finally, the remaining time is estimated during run-time using the calculated process execution speed.

We tested our model with two datasets: a medical dataset containing depth video and audio records of 35 actual trauma resuscitations at Children's National Medical Center (CNMC) \cite{li2016online}, and an Olympic swimming dataset containing 60 YouTube video records of Olympic swimming competitions in different styles. For the resuscitation dataset, our process progress estimation system achieved 86\% average online phase detection accuracy and 0.67 F1-score, outperforming existing systems. Specifically, our system incurred 12.65\% completeness estimation error, with an average 7.5 minute remaining-time estimation error (14\% of total duration). For the Olympic swimming dataset, the system achieved 88\% accuracy with 0.58 F1-score and incurred 6.32\% completeness estimation error with average 2.9 minute remaining-time estimation error (18\% total duration). Using additional network layers to replace the GMM, our system slightly outperformed existing process-phase detection systems on the trauma resuscitation dataset \cite{twinanda2017endonet,stein2013combining}, while simultaneously providing progress and the remaining time estimation. Our phase-detection system for trauma resuscitation has been deployed in a trauma room and is currently generating data for future applications. Our contributions are:

\begin{enumerate}
	\item A novel deep regression-based approach for process progress estimation and phase detection using commercially available sensors, as well as a new rectified hyperbolic tangent (\textit{rtanh}) activation function that bounds the regression value to a meaningful range and accelerates the neural network training.
	
	\item A GMM-based phase detection approach based on completeness regression results, as well as a conditional loss function for model tuning using regression and classification error.
	
	\item A deep learning structure that models the progress of \textit{nonlinear} processes, which is derived from our structure for \textit{linear} process modeling.
	
	\item The trauma resuscitation dataset and Olympic swimming dataset that can be used in future research, as well as the details of our system deployment in an actual medical setting (can be used as a reference for other applications).
\end{enumerate}

The rest of the paper is organized as follows: Section \ref{sec:02} introduces the problem. Section \ref{sec:03} presents our proposed method. Section \ref{sec:04} describes our experimental results and performance comparison to other systems. Section \ref{sec:05} discusses the results, and Section \ref{sec:06} concludes the paper.

\section{Related Work}
\label{sec:02}

Process progress modeling can be done from three aspects: the completeness, the process phase, and the remaining time. Most of the previous research only focused on the process phase detection. Based on different application scenarios, previous works approached process modeling in three ways. 

Systems designed to focus only on process modeling or work in very specific applications (such as certain types of surgery) used  manually generated event logs \cite{forestier2015automatic} or the medical equipment signals \cite{padoy2012statistical} as the input. These types of input data are relatively noise free, and can be directly used for process modeling without preprocessing and feature extraction. Previous approaches treated process phase detection as a classification problem and directly applied shallow classifiers commonly used for sequential data analysis (e.g., HMM or decision tree \cite{padoy2012statistical}). Such approaches have proven accurate and easy to implement, and some association rules could be mined by analyzing the generated HMM transition matrix or decision tree topology \cite{herbst1998integrating}. But the drawback of these approaches is the use of manually generated event log or the data from specific equipment, making these systems hard to deploy, and the trained classifier may have difficulty working with high-dimensional and noisy sensor data.

To avoid such limitations, we designed our system to use commercially available sensors as the input source. There are previously proposed process phase detection systems using different sensor data \cite{blum2010modeling,li2016deep,li2016online}. For daily living and certain surgery scenarios, the RGB camera and wearable sensors were used to capture the gesture and posture for process phase estimation \cite{blum2010modeling}. For more constrained scenarios, such as in medical settings, the less intrusive and privacy-preserving sensors such as passive RFID or depth camera were used \cite{li2016deep,li2016online,bardram2011phase}. Because the sensor data collected in a real-world environments is noisy (may contain hardware noise, outliers, and missing data), machine learning algorithms were commonly used to establish the connection between sensor data and the process phase. On the other hand, the feature and classifier selection were often chosen arbitrarily or empirically \cite{li2016deep,li2016activity}, which makes the shallow-modeled systems difficult to transfer. In addition, when processing multi-modal input data, the shallow-modeled systems usually make classification independently based on each input modality and combine them by voting \cite{stikic2008adl}. In this way, potential correlations between different input modalities are ignored.

This limitation of shallow classifiers is not unique for process progress modeling; many fundamental computer science fields (such as image classification and speech recognition) face the same issue. In recent years, the CNN and LSTM \cite{simonyan2014very,hochreiter1997long} have been successfully implemented in these fields and achieved significantly better performance compared with the state-of-the-art shallow-modeled solutions \cite{lecun2015deep}. The CNN was widely used in image classification and image feature extraction, because the CNN with the learnable filters can automatically learn the representative spatial features from the raw training data and is proven generalizable (the pre-trained model can be used as the feature extractor) \cite{girshick2015fast}. The current state-of-the-art process progress estimation systems use the pre-trained AlexNet \cite{twinanda2017endonet} for process classification using video frames as input, but this pure CNN does not consider temporal associations between features in adjacent frames. To compensate for lost temporal associations that maintain the logical phase order, previous research attempted using a time window \cite{stauder2016tum}, HMM\cite{twinanda2017endonet}, or modified softmax layer \cite{li2016online}. However, the HMM does not scale well to complex process with a large number of phases, because the topology of HMM's transition matrix becomes less representative and requires manual tuning. The time-window approach prevents the system from working online. Based on the LSTM's ability to model temporal associations in speech recognition \cite{graves2013speech} and natural language understanding \cite{xu2015show}, we implemented the LSTM with CNN in our system to extract the spatio-temporal features. We then introduced a regression model with GMM to achieve both process completeness estimation and phase detection.

\section{System Structure}
\label{sec:03}

Our system consists of four main parts (Fig. \ref{fig:01}): 

\begin{figure}[!htb]
	\centering
	\includegraphics[width=0.75\textwidth]{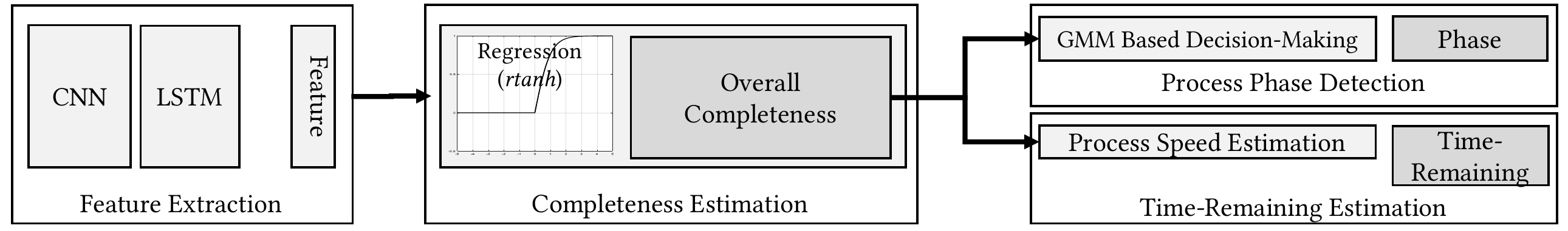}
	\captionsetup{justification=centering}
	\caption{The overall system structure used for modeling process progress estimation.}
	\label{fig:01}
\end{figure}

\textbf{Feature extraction:} Because we used different sensors for data collection in different applications, the first step is the data representation and feature extraction. Instead of using manually crafted features, we used a CNN and LSTM based model to learn the spatio-temporal features from the input data \cite{sainath2015convolutional}. We implemented a multimodal structure to fuse the features extracted from different sensor data.

\textbf{Completeness estimation:} The system has a deep regression model that directly produces a single regression output value. We introduced the \textit{rtanh} activation function for bounding the neuron output to a valid range.

\textbf{Phase detection:} The system takes estimated completeness as input and uses a probabilistic GMM inference to detect the process phase. Phase detection provides a conditional loss function to help train the regression model.

\textbf{Remaining-time estimation:} Our system dynamically updates the estimation of remaining time based on the observed speed of process execution, which we defined as the rate at which one percent of the process is being accomplished.

\subsection{Feature Extraction}
\label{sec:031}

Similar to the activity recognition \cite{karpathy2014large,li2016deep}, our estimation of the process progress relies on both spatial and temporal features. The spatial features in a video frame define the activity at a time instance, while the temporal features from consecutive frames define a phase in the process. The learnable filters in CNNs are commonly used to extract the spatial features \cite{zeiler2014visualizing}, and LSTMs are often used to model temporal dependencies of the sequential data \cite{karpathy2015visualizing}.

\begin{figure}[!htb]
	\centering
	\includegraphics[width=0.75\textwidth]{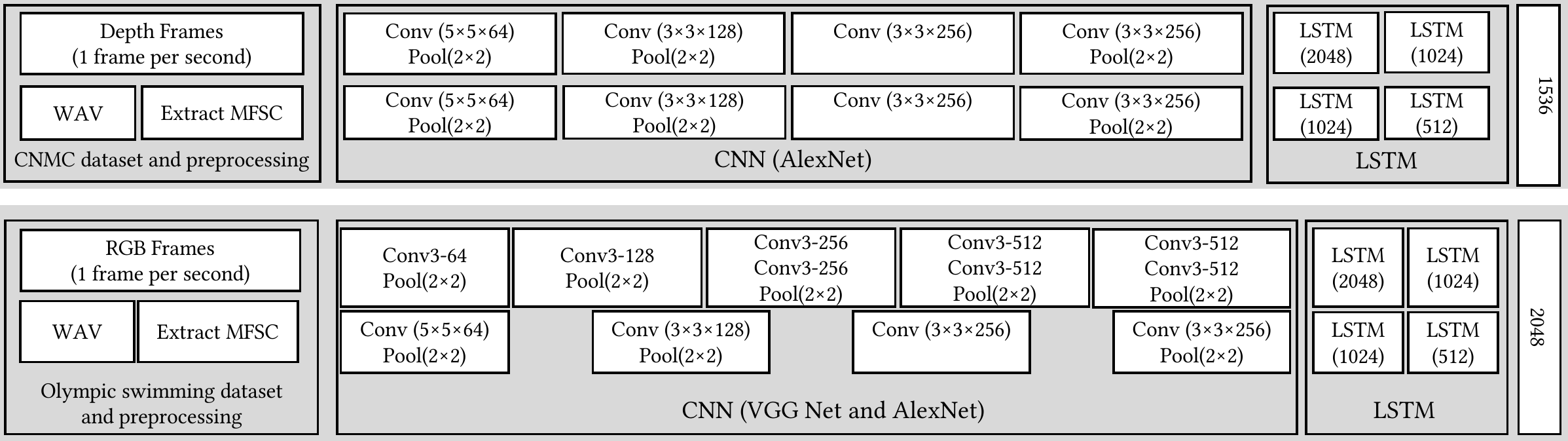}
	\captionsetup{justification=centering}
	\caption{Feature extraction framework for the trauma resuscitation dataset (top) and the Olympic swimming dataset (bottom).}
	\label{fig:02}
\end{figure}

Our trauma resuscitation dataset contains low-resolution depth images and audio from a Kinect \cite{li2016online}. We chose the Kinect depth sensor for our medical application because it is privacy-preserving (does not capture any facial details). Other types of sensors can be easily combined by introducing additional multimodal branches into our system. During every second, the video input branch was directly fed depth frames, while the audio branch was fed MFSC feature maps \cite{abdel2014convolutional} for feature extraction. A CNN-LSTM structure then performed spatio-temporal feature extraction, where we used a pre-trained CNN structure (AlexNet \cite{krizhevsky2012imagenet}), followed by multiple LSTM layers. The features extracted from different sensors are subsequently combined in a fusion layer, following our previous implementation \cite{li2016online} (Fig. \ref{fig:02}, top).

The Olympic swimming dataset was collected from YouTube videos recorded by different types of cameras (including cell phone cameras and professional cameras). This dataset has videos of different resolutions and audio recorded by cameras at different distances. Due to our hardware limitation (described in Section 3.5 in Section 3.5), we downsampled the video frames from 25 fps to 10 fps and resized them to $256\times256px$. Instead of AlexNet, we used a deeper VGG Net (Fig. \ref{fig:02}, bottom) with pre-trained weights \cite{simonyan2014very} for image feature extraction (Fig. \ref{fig:02}), because RGB images contain more textural details than depth images.

\subsection{Process Regression and Completeness Estimation}
\label{sec:032}

\begin{figure}[!htb]
	\centering
	\includegraphics[width=0.7\textwidth]{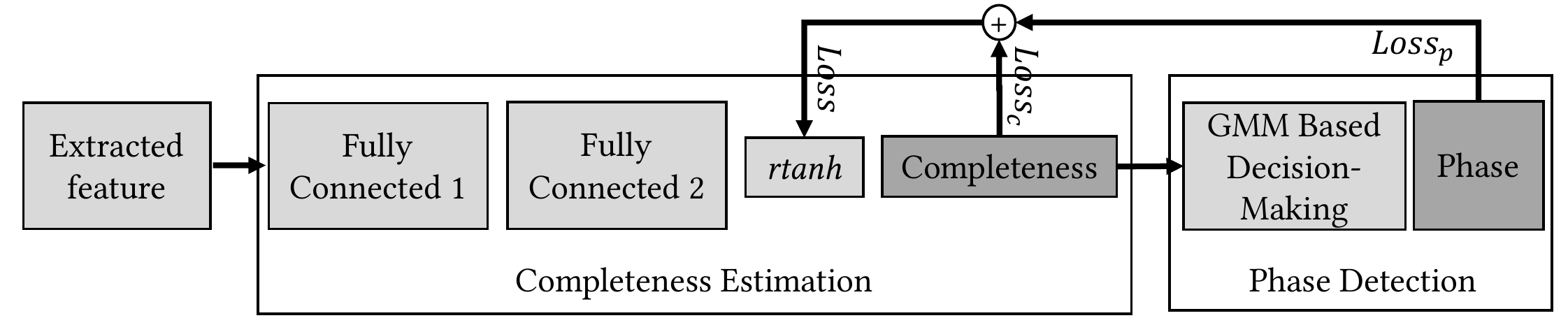}
	\captionsetup{justification=centering}
	\caption{The system diagram for completeness estimation and phase prediction. The $Loss_c$ represents the loss from completeness regression error and the $Loss_p$ represents the loss from phase prediction error.}
	\label{fig:03}
\end{figure}

Unlike image classification or single-image activity recognition where each sample corresponds to an individual image, process completeness is a continuous variable which cannot be estimated as a discrete classification problem. For this reason, we introduce a regression-fitting model that uses the extracted spatio-temporal features for process completeness estimation (Fig. \ref{fig:03}, completeness estimation part).  

Our deep neural network extracts features from input data and performs regression using an activation function which generates the regression value $\hat{y}\in[0,1]$. The activation of regression neuron can be expressed as:
\begin{equation}
	\label{eqn:01}
	\hat{y}=f(\boldsymbol{\omega}^T\boldsymbol{\phi}+b)
\end{equation}
where $\boldsymbol{\omega}$ $(N\times1)$ is the weight vector and $\boldsymbol{\phi}$ $(N\times1)$ is the output of the last fully connected layer (Fig. \ref{fig:03}, fully connected layer 2, with $N$ neurons in total), $b$ is the bias term and $f(\cdot)$ is the activation function of the output neuron. To model the process progress, the activation function must have the following properties: (1)~ the activation output should be zero if the process has not yet started (e.g. pre-arrival phase of the trauma resuscitation);  (2) the regression value changes continuously from zero to one as the process proceeds; and (3)~ the activation should be equal or close to one, with a proper strategy (e.g., thresholding) for when the process has been completed (e.g., after the patient left the trauma room).  

We modified the \textit{hyperbolic tangent function} for this purpose, and we named it the rectified hyperbolic tangent (\textit{rtanh}) function:
\begin{equation}
	\label{eqn:02}
	rtanh(x)=\max(0, \tanh(x))=\max(0, \frac{e^{2x}-1}{e^{2x}+1})
\end{equation}

This activation function returns zero for negative inputs and positive values up to one for positive inputs. This value range makes it suitable for returning a progress completeness percentage ranging from 0\% to 100\%. The \textit{sigmoid} function has the same value range, but its range of positive values is distributed across all real numbers, making it slower to train. In addition, during the backpropagation, the gradient of \textit{rtanh} activation can better prevent gradient vanishing than the \textit{sigmoid} because the derivative of $tanh\le1$ and of $sigmoid\le0.25$ \cite{hochreiter1998vanishing}. This property will lead to a faster training under the same setting of the learning rate. To compare the performance of the \textit{rtanh} and \textit{sigmoid} functions, we initialized the neural network parameters to the same value and trained the network using \textit{rtanh} and \textit{sigmoid} with the same training and testing split of the data. Our experiments showed that the two activations eventually led to similar accuracies, but the convergence time (the time it takes until accuracy stabilizes for 3 epochs) using \textit{rtanh} was 30\% shorter than that using \textit{sigmoid}, because of \textit{rtanh}'s steeper gradient. As we focused on modeling a linear process, the current completeness value should be increasing compared to the previous completeness value. We used an LSTM after the CNN to retain the estimated progress information and help the system with tracking the increasing completeness trend.

Similar to a classifier, a regression model can be trained using backpropagation. When generating the ground truth label, the completeness is labeled differently depending on the application. For example, the completeness of a trauma resuscitation equals zero before the patient entered the room and remains at one after the patient left. Given the time at which the data were collected relative to the beginning and end of a process enactment, we can label the data from a certain time instance with the completeness range to which this time instance belonged. We divided the data into twenty 5\%-segments, and labeled each segment with an associated completeness between 0\% (process start) and 100\% (process end), forming a stepwise function with 5\% increments. The output of the \textit{rtanh} neuron can be directly used as the regression result (Fig. \ref{fig:03}, decision-making part). 

To make the regression smooth as the completeness progresses for real world processes, we applied a Gaussian smooth filter after the \textit{rtanh} neuron. The loss from completeness estimation error $Loss_c$ can be expressed as: 
\begin{equation}
	\label{eqn:03}
	Loss_c(\theta=\{w,b\},D)=\frac{\sum_{i=0}^{|D|}abs(R(\theta=\{w,b\},D_i)-p_i)}{|D|}
\end{equation}
where the $\theta=\{w,b\}$ denotes the model with parameter set $\theta$, and $D$ is the input dataset. $R(\theta=\{w,b\},D_i)$ denotes the regression output from the model with parameter set $\theta$ on dataset $D$ at time instance $i$ as input. $p_i$ denotes the $i^{th}$ percentage label for regression in terms of process completion. We used the mean absolute error for the loss, which is the $abs(\cdot)$ term in the loss function. Although other error measurements such as mean square error could be used, the square error would inhibit training by making the losses even smaller (considering that $errors\in[0,1]$).

\subsection{Phase Detection and Conditional Loss}
\label{sec:033}

With the completeness estimated by the regression model, we need to predict the discrete phase based on the continuous completeness estimation results. A separate classification model using the same extracted features can be used for phase detection \cite{li2016online}, but performing classification without considering the completeness ignores the temporal associations between the extracted process progress features. 

\begin{figure}[!htb]
	\centering
	\includegraphics[width=0.8\textwidth]{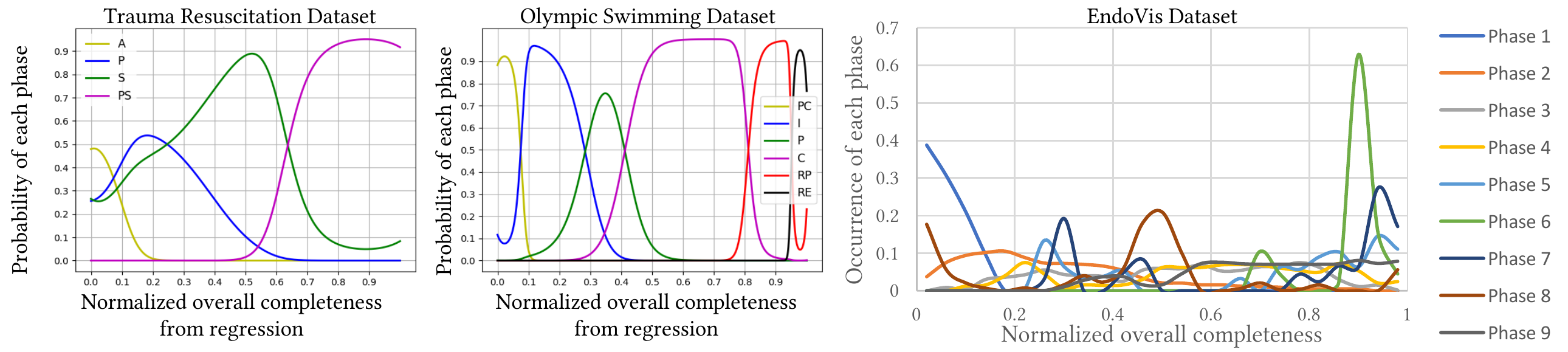}
	\captionsetup{justification=centering}
	\caption{The probability for each phase plotted against the normalized process duration for the trauma resuscitation dataset (left) and the Olympic swimming dataset (middle) and EndoVis dataset (right) that contains nonlinear process.}
	\label{fig:04}
\end{figure}

Given that the processes we considered are linear, we found that the duration of each phase is similar to the Gaussian distribution. Therefore, we assumed a Gaussian distribution for the duration of each phase, which allows us to use a Gaussian mixture model (GMM) with one centroid per phase for phase prediction (Fig. \ref{fig:04}, left \& middle). On the other hand, this GMM framework does not generalize to nonlinear processes with duplicate or rearrangeable phases (Fig. \ref{fig:04}, right EndoVis dataset). Nonlinear processes require modifications in the GMM-based model as discussed in Section \ref{sec:05}.

Training the GMM \cite{bilmes1998gentle} only requires the occurrence distribution for each phase on the normalized process duration scale, which can be obtained from the ground truth data. We can pre-train the GMM using completeness ground truth and establish the probabilistic association between the overall completeness and process phase. The phase recognition results can be calculated from:
\begin{equation}
	\label{eqn:04}
	\hat{p}={\mathrm{argmax}}_{1\le{k}\le{K}}
	\{\log(w_k)-\frac{1}{2}\log(\det(2\pi \boldsymbol{\Sigma}_k))-\frac{1}{2}(\boldsymbol{x}-\boldsymbol{\mu}_k)^T\boldsymbol{\Sigma}_k^{-1}(\boldsymbol{x}-\boldsymbol{\mu}_k)\}
\end{equation}
where $\hat{p}$ is the predicted phase. $\boldsymbol{x}$ is the completeness estimated by the regression model. $w_k$ is the weight for the $k^{th}$ Gaussian kernel. $\boldsymbol{\mu}_k$ and $\boldsymbol{\Sigma}_k$ denote the mean vector and the covariance matrix for the $k^{th}$ Gaussian kernel. $K$ is the total number of phases. ${\mathrm{argmax}}_{1\le{k}\le{K}}\{\cdot\}$ denotes the function finding the index $k$ with the largest likelihood among all $K$ indices.

With equation \ref{eqn:04} and the regression result, we can directly use the GMM for phase prediction (Fig. \ref{fig:04}). We only applied the GMM for four out of the six phases in the trauma resuscitation dataset, removing the pre-arrival and patient-leave phases. This is because the duration of the pre-arrival phase depends on the transport time from the injury scene to the hospital, which was not recorded in the dataset; and the patient-leave phase is coded as an ending signal lasting only one second. Our GMM-based phase prediction approach relies only on the regression result, and the phase prediction error does not impact the backpropagation training process used to tune the regression. The regression model can be trained based on both the completeness estimation error and the phase prediction error. Based on the assumption that the regression value is correct, the system generates no additional loss if the GMM-based phase detection result is correct. Otherwise, an additional loss is incurred, calculated as the distance from the regression result to the mean of occurrence distribution for the actual current phase. The loss from phase estimation error $Loss_p$ is conditional based on the phase detection result and can be expressed as:

\begin{equation}
	\label{eqn:05}
	Loss_p=
	\begin{cases}
		0       & \quad \hat{p}=p\\
		|R(D_p)-\mu_p|  & \quad \hat{p}\neq{p}\\
	\end{cases}
\end{equation}
where the $\hat{p}$ denotes the GMM-predicted phase and $p$ is the actual current phase. $D_p$ is the input data for phase $p$ and $R(D_p)$ denotes the regression model output. The $\mu_p$ is the mean of the Gaussian distribution for the actual current phase $p$. By combining the loss from regression error and the classification error, the regression model can be tuned to make completeness estimations and phase predictions that obey a logical order of phases. During the training, we added weights $\alpha$ for $Loss_c$ and $\beta$ for $Loss_p$ in equation \ref{eqn:06}:
\begin{equation}
	\label{eqn:06}
	Loss=\alpha{Loss_c}+\beta{Loss_p}
\end{equation}

\begin{figure}[!t]
	\centering
	\includegraphics[width=0.75\textwidth]{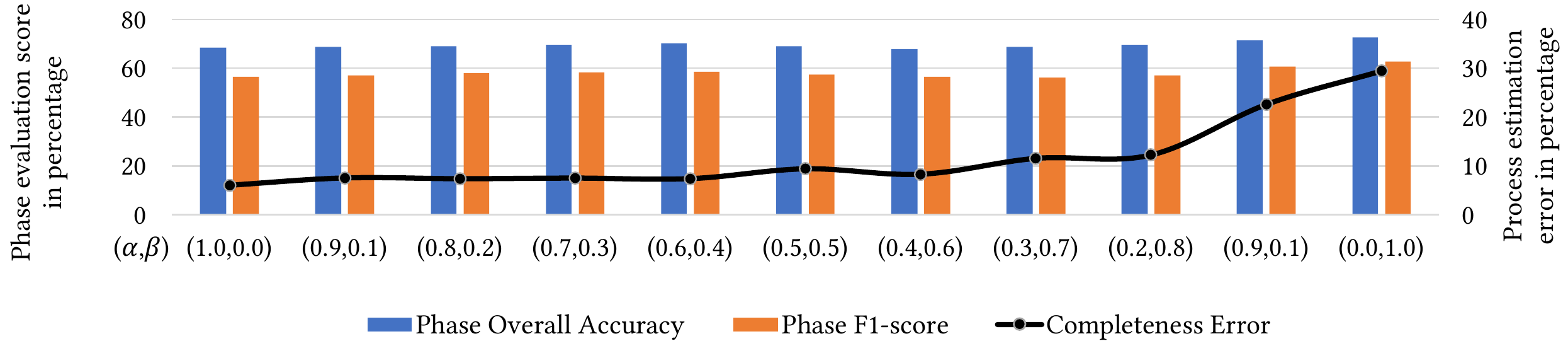}
	\captionsetup{justification=centering}
	\caption{The process completeness estimation error and phase detection accuracy with F1-score using different $\alpha,\beta$ values.}
	\label{fig:05}
\end{figure}

When $\alpha=1$ and $\beta=0$, the model becomes a regressor trained only on the regression error. Alternatively, when $\alpha=0$ and $\beta=1$, the model becomes a classifier trained only on the classification error. Training the system with a larger $\alpha$ would cause the system to prioritize overall completeness regression instead of minimizing phase prediction error, and a larger $\beta$ would do the opposite. We determined $\alpha,\beta$ empirically by minimizing the completeness estimation and phase classification error based on a small subset of each dataset. We used 10 training cases from both datasets to determine the ratio of $\alpha$ vs. $\beta$. Fig. \ref{fig:05} shows the averaged completeness estimation error and process phase detection accuracy with F1-score on the Olympic swimming dataset. We selected $\alpha=0.6$ and $\beta=0.4$ which achieved the best balance between the completeness and phase detection.

\subsection{The Remaining Time Estimation}
\label{sec:034}

The speed of process performance may change, which means that our remaining-time estimator requires a dynamically-updated estimation strategy. We estimated the remaining time by calculating the average speed of completing 1\% of the process. If $\rho$ represents the current completeness (in percentage) and $\tau$ is the time elapsed since the start, the remaining time $t$ can be estimated as:
\begin{equation}
	\label{eqn:07}
	t=(\tau/\rho)\times(1-\rho)
\end{equation}
where $(\tau/\rho)$ denotes the time needed to make 1\% overall progress and $(1-\rho)$ denotes the remaining percentage for the process to finish. During the runtime, we updated the remaining-time estimation for every second.

\subsection{Implementation}
\label{sec:035}

We implemented our model with the \textit{Keras} framework using \textit{TensorFlow} backend. We implemented our \textit{rtanh} activation and loss functions using the tensor interfaces of the \textit{Keras}. As proposed in previous research \cite{krizhevsky2012imagenet}, we used the rectified linear unit (ReLU) as our CNN activation function. For the trauma resuscitation dataset, we initialized and trained the model using a single GTX 1080 GPU. For the YouTube Olympic swimming dataset that has a larger input frame resolution and larger network structure, we used dual GTX 1080 GPUs for training. We used the weights trained in VGG Net \cite{simonyan2014very} to initialize the VGG Net for RGB frame processing (Fig. \ref{fig:02}, bottom). The Adam optimizer \cite{kingma2014adam} was implemented with initial learning rate of 0.001 and a decay of $10^{-8}$. We configured the system to stop automatically if the performance did not change for three consecutive epochs using the \textit{Keras} callback function.

Due to the large model size, we adopted the dropout strategy \cite{srivastava2014dropout} in the network to avoid overfitting. We also partitioned the training and testing sets by whole case instead of segmenting each case for evaluation, to minimize similarities between training and testing sets, as suggested in \cite{li2016online}.

Unlike image classification models with a target in each training image, process phase classification may be significantly different from case to case (e.g., treating patients with different injuries or performing different swimming styles). Entering data into the classifier simultaneously causes slow convergence. For this reason, we used a training strategy: we initially fed only two process enactment cases into the classifier. When the system achieved a specified loss value (manually defined), we fed one more case into the classifier and kept feeding in new cases after each step until all cases have been used. Using this approach, the model learned the specific scenarios rapidly and later was able to discriminate between similar classes in other cases. 

\section{Experimental Results}
\label{sec:04}

\subsection{Data Collection}
\label{sec:041}

\begin{figure}[!htb]
	\centering
	\includegraphics[width=0.7\textwidth]{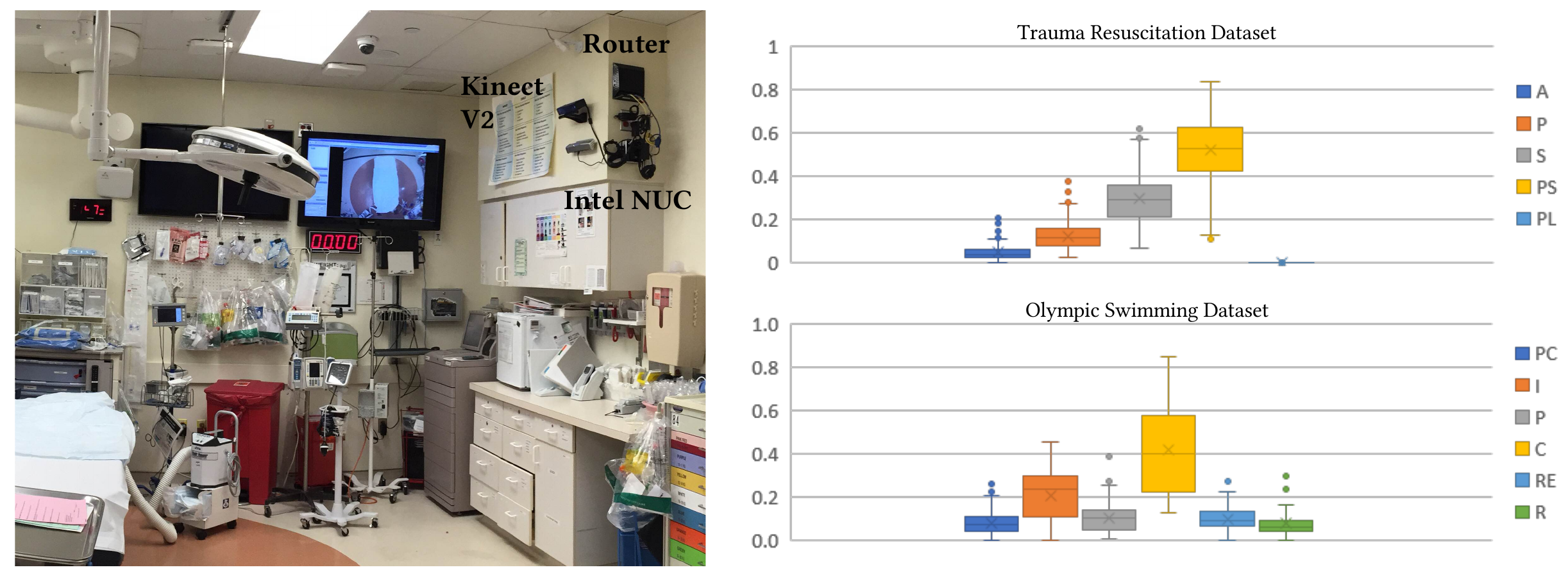}
	\captionsetup{justification=centering}
	\caption{The trauma room with our Kinect installed (left) and the boxplot of phase duration in completeness percentage for trauma resuscitation dataset and Olympic swimming dataset (right).}
	\label{fig:06}
\end{figure}

\textbf{Trauma Resuscitation Dataset:} Our trauma resuscitation dataset was collected in a trauma room at the Children's National Medical Center in Washington D.C. Use of this data and its related research have been approved by the hospital's IRB. 150 trauma resuscitation cases were manually coded as the ground truth. The data was collected through a Kinect depth sensor mounted on the side wall of the trauma room \cite{li2016deep,li2016online} (Fig. \ref{fig:06}, left). Of the 150 trauma resuscitation cases, 50 cases had synchronized depth data and 35 from these 50 cases also had the synchronized audio data. We used the 150 coded cases to generate the GMM phase distributions (Fig. \ref{fig:04}), and 35 cases with both depth video and audio data for model training and testing. Given the different patient conditions and times of the day, the durations of resuscitation phases varied (Fig. \ref{fig:06}, right). The system recorded depth video at 1 fps to save storage space. \\
\textbf{Olympic Swimming Dataset:} The Olympic swimming dataset includes 60 videos from 2004 to 2016 downloaded from YouTube. The videos were recorded by different devices and at different angles and distances. We coded the ground truth for the six phases of swimming competitions manually. We manually edited some of the downloaded videos to ensure that all the video clips contained only swimming-related content. Because some videos did not record the pre-competition phase or result phase, these cases did not have all six phases. 

\subsection{Evaluation of Process Progress Estimation}
\label{sec:042}

We evaluated the proposed system with the trauma resuscitation dataset and the Olympic swimming dataset for estimating overall completeness, phase, and the remaining time. Because the dataset is imbalanced, we used the weighted average for all evaluations.

\subsubsection{Completeness Estimation}
\label{sec:0421}

\begin{figure}[!b]
	\centering
	\includegraphics[width=0.75\textwidth]{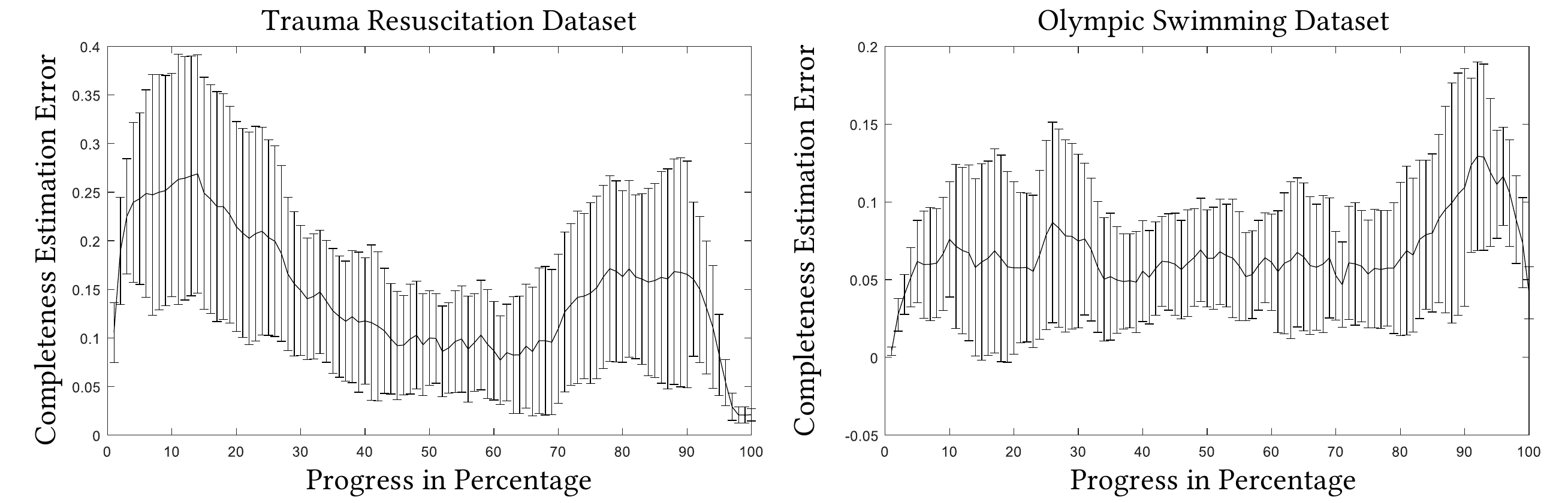}
	\captionsetup{justification=centering}
	\caption{Mean (solid line) and Variance (shaded region) of completeness estimation error plotted against normalized duration of process enactment.}
	\label{fig:07}
\end{figure}

\begin{figure}[!htb]
	\centering
	\includegraphics[width=0.9\textwidth]{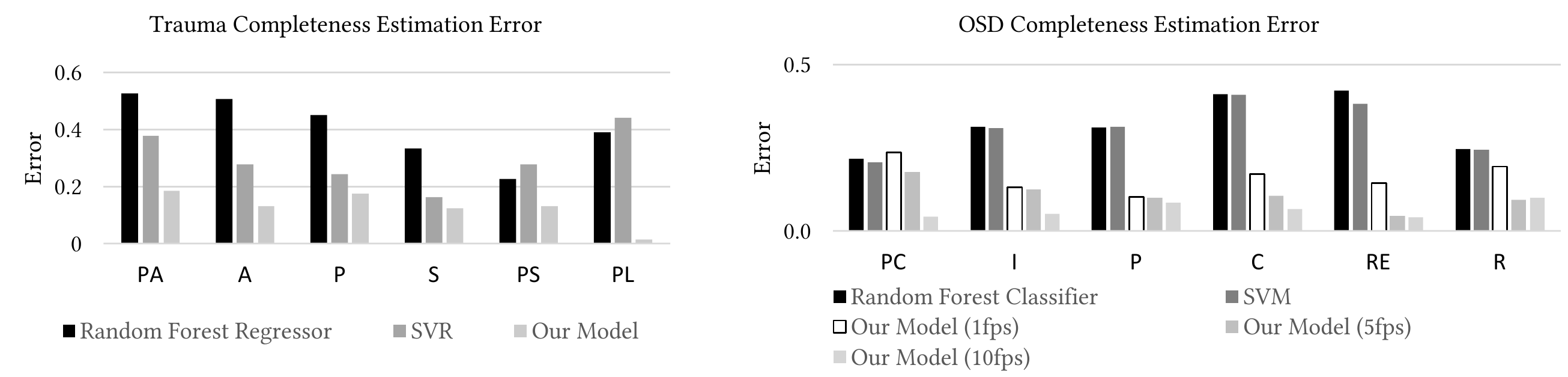}
	\captionsetup{justification=centering}
	\caption{Aggregate error over different phases of resuscitation and swimming processes.}
	\label{fig:08}
\end{figure}

We first calculated the overall completeness estimation error (Mean Absolute Error (MAE)) of the system by inputting 20\% of cases into the trained network. Our system achieved an average 12.65\% overall completeness error for trauma resuscitations and 6.32\% error for Olympic swimming dataset. For the MAE of the completeness estimation with normalized process duration of testing cases (Fig. \ref{fig:07}), a large MAE indicates that the system had difficulty distinguishing certain scenarios and a large variance indicates that the system did not generalize well for the testing cases. We further evaluated the completeness estimation error on each process phase (Fig. \ref{fig:08}), and showed that the system performed well in the starting and ending phases, which is due to the ability of our proposed \textit{rtanh} function to hold regression output at zero (starting) and one (ending). The high error rate in the post-secondary phase may be related to the high similarity in appearance between the activities in secondary and post-secondary phases. Different types of sensors could be introduced to improve process progress estimation. The competition and replay phases of the Olympic swimming dataset had a higher error rate because replay phase is a slow-motion version of the competition phase and at low frame rates (e.g. 1 fps), the two phases are barely distinguishable. In addition, having higher-frame-rate videos containing more information can reduce overfitting  \cite{srivastava2014dropout}. To confirm this, we sampled the video at  1, 5 and 10 fps to retrain the system using the same setting of the parameters. The results showed that a higher frame rate significantly decreases completeness estimation error during the replay phase (Fig. \ref{fig:08}). In the rest of this paper, we used 10 fps for the Olympic swimming dataset and 1 fps for the trauma resuscitation dataset, unless specified otherwise.

These results suggest that the performance of progress completeness estimation can be further improved by
(1)~ Adding other sensors to provide more input features and better distinguish the phases with similar sensor data. 
(2) Using a higher frame rate for data preprocessing to retain more temporal associations and prevent overfitting. As a consequence, the computational resources and storage space cost would increase. Based on our experiments, a 5-minute video sampled at 10 fps with $256\times256px$ resolution takes more than 8GB of storage;
(3) Using deeper models to extract more abstract features. As shown in computer vision research, using ResNet \cite{he2016deep} instead of VGG may lead to better performance. 

We performed additional experiments to confirm that our system structure is superior to the shallow models. We used the common shallow regressors of Support Vector Machine (SVM) and Random Forest and we trained them with the same raw video and audio data used for our deep model. Our deep model achieved significantly better accuracy than the shallow models, confirming that the CNN-LSTM extracts better multimodal features (Fig. \ref{fig:08}). The shallow models faced difficulties extracting features from the raw input, and manually crafted features would be necessary to enhance their performance.

\subsubsection{Phase Detection}
\label{sec:0422}

\begin{figure}[!t]
	\centering
	\includegraphics[width=0.6\textwidth]{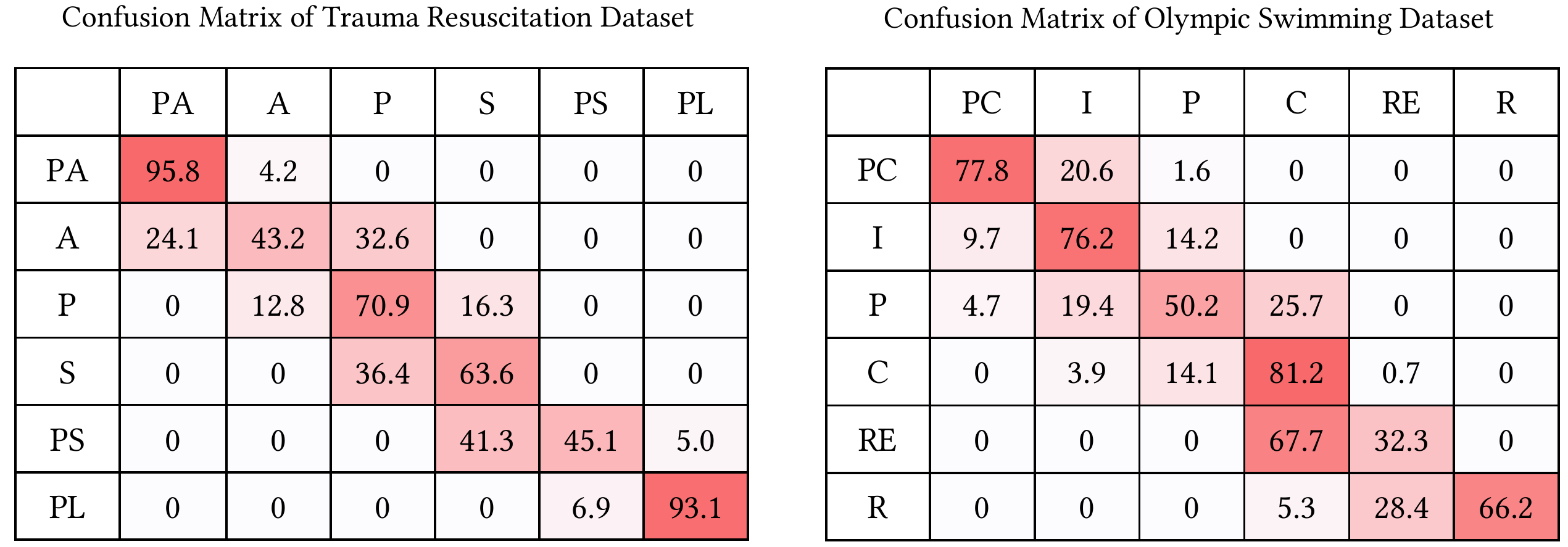}
	\captionsetup{justification=centering}
	\caption{The confusion matrices for predicting phases of trauma resuscitation and swimming competition.}
	\label{fig:09}
\end{figure}

To evaluate the performance of process phase detection, we first generated the confusion matrices of the phase prediction results (Fig. \ref{fig:09}). Our system achieved an average 86.06\% phase prediction accuracy for trauma resuscitation data and 87.99\% for the Olympic swimming data. Our analysis of the confusion matrices showed that our system is able to make logical phase predictions (Fig. \ref{fig:09}). The zero values in most of the upper and lower triangles of the confusion matrices indicate that the system predictions rarely jumped between non-adjacent phases. This observation confirmed that our activation function helps maintain the estimated completeness range, and the LSTM enforces an ascending overall completeness prediction. By further comparing the confusion matrices of both datasets, we found that: 
(1) The system accurately predicted the starting and ending phases, due to the \textit{rtanh}'s ability to maintain the zeros and ones for the associated starting and ending phases. 
(2) The phase detection performance is associated with the regression performance. For example, the regression and phase detection systems both poorly predicted the post-secondary phase compared to other phases of trauma resuscitation (Fig. \ref{fig:08}). This is because the GMM took the process completeness estimation results as the input for process phase detection, and the error made in the process completeness estimation was propagated to the process phase detection. 
(3) The system achieved a similar performance for both datasets, despite the differences in camera mobility. The trauma resuscitation dataset was recorded by a fixed camera, while the Olympic swimming dataset was recorded by moving cameras. The fixed camera captured the activity's scene and provided continuous activity sequence information, while the moving camera might capture some discontinuous unrelated data. Our approach achieved the similar performance under both scenarios, showing its potential for analyzing the temporal activity processes regardless of the video input's changing viewpoints, due to the ability of CNN-LSTM structure to learn and extract the representative features.  

We then analyzed how our model outperformed the shallow models in learning the features automatically. Because process phases generally have different durations and the amount of available data is imbalanced for different phases, we used three different metrics to perform a comprehensive evaluation. The F-measure (weighted average scores for precision, recall, and F-score) and 2SET metrics \cite{ward2011performance} were used to break down the wrongly classified instances into fragmentation, over-filling, and under-filling \cite{ward2011performance} (Table \ref{tab:02}). Because the transition between process phases is instantaneous, the insertion, deletion, and merging metrics do not apply to our datasets \cite{powers2011evaluation}. To compensate for the F-measure's disregard for false positive samples, we used informedness, markdness, Matthews correlation coefficient (MCC) for a more comprehensive evaluation (Table \ref{tab:02}).

\begin{table}[!b]
	\centering
	\setlength{\belowrulesep}{0pt}
	\setlength{\aboverulesep}{0pt}
	\captionsetup{justification=centering}
	\caption{Phase prediction performance comparison using different modalities. The shallow classifier based results are shaded.}
	\label{tab:02}
	\begin{minipage}{\columnwidth}
		\begin{center}
			\begin{tabular}{|p{0.28\textwidth}|p{0.055\textwidth}p{0.055\textwidth}p{0.055\textwidth}|p{0.055\textwidth}p{0.055\textwidth}p{0.055\textwidth}|p{0.055\textwidth}p{0.055\textwidth}p{0.055\textwidth}|}
				\toprule
				\textbf{Input Data Source} & \textbf{Prec.} & \textbf{Rec.} & \textbf{F1-s.} & \textbf{Info.} & \textbf{Mark.}& \textbf{MCC} & \textbf{Frag.}&\textbf{Under.} & \textbf{Over.}\\ \midrule
				Trauma audio&0.33&0.37&0.17&0.09&0.06&0.07&0.01&0.13&0.13\\ \midrule
				Trauma video&0.63&0.57&0.51&0.35&0.43&0.39&0.01&0.10&0.15\\ \midrule
				\textbf{Trauma audio \& video}&\textbf{0.76}&\textbf{0.68}&\textbf{0.67}&\textbf{0.35}&\textbf{0.54}&\textbf{0.54}&\textbf{0.00}&\textbf{0.08}&\textbf{0.12}\\ \midrule			
				Swim audio&0.41&0.40&0.32&0.29&0.29&0.26&0.00&0.11&0.23\\ \midrule
				Swim video&0.58&0.55&0.48&0.37&0.43&0.39&0.00&0.09&0.24\\ \midrule
				\textbf{Swim audio \& video} &\textbf{0.67}&\textbf{0.70}&\textbf{0.58}&\textbf{0.52}&\textbf{0.53}&\textbf{0.48}&\textbf{0.00}&\textbf{0.06}&\textbf{0.12}\\ \midrule
				Trauma audio \& video (SVM)&0.09&0.37&0.09&0.07&0.03&0.07&0.03&0.01&0.13\\ \midrule
				Trauma audio \& video (SVM)*&0.46&0.22&0.30&0.16&0.05&0.06&0.02&0.16&0.06\\ \midrule
				Trauma audio \& video (RF)&0.17&0.25&0.16&0.01&0.06&0.01&0.08&0.17&0.01\\ \midrule
				Trauma audio \& video (RF)*&0.44&0.29&0.34&0.12&0.08&0.03&0.02&0.16&0.02\\ \midrule
				Swim audio \& video (SVM)&0.59&0.24&0.24&0.07&0.18&0.17&0.00&0.15&0.16\\ \midrule
				Swim audio \& video (SVM)*&0.61&0.27&0.27&0.13&0.30&0.31&0.00&0.15&0.15\\ \midrule
				Swim audio \& video (RF)&0.34&0.22&0.14&0.01&0.08&0.04&0.07&0.14&0.04\\ \midrule
				Swim audio \& video (RF)*&0.46&0.30&0.21&0.03&0.14&0.07&0.00&0.16&0.19\\
				\bottomrule
				\multicolumn{10}{@{}l}{*\footnotesize{classifier takes the output of last fully connected layer as input.}}
			\end{tabular}
		\end{center}
	\end{minipage}
\end{table}

As previously mentioned, shallow-modeled regressor generated significantly higher progress completeness estimation errors (Fig. \ref{fig:08}), which propagated to the phase detection. To focus on the classification comparison, we built shallow phase classifiers (SVM and Random Forest) using both raw input data and CNN-extracted features (Fig. \ref{fig:03}, fully connected layer 2). The comparison of the metrics (Table \ref{tab:02}) revealed that 
(1) Our deep model was able to achieve significantly higher precision and recall; most predictions were correct and most true instances were detected.
(2) Our deep model achieved higher MCC score, indicating a positive association between the prediction and the ground truth. 
(3) Our deep model generated predictions with almost no fragmentation, while the predictions by the shallow classifier sometimes had fragmentations. 
Considering that there were gaps between the phases, the zero-fragmentation indicated that our system was able to make logically-ordered phase detection due to the use of LSTM to model temporal associations of features. Again, since we were calculating the 2SET scores using one-versus-rest method, the over-filling of one phase would result in under-filling of an adjacent phase. The over-filling and under-filling were similar to each other (not equal because we were calculating their weighted averages). Our deep model achieved lower under-filling and over-filling, demonstrating the ability to detect the phase transition points more accurately.  

Since we used a multimodal structure, we further studied the impact of using single and both modalities. The results (Table \ref{tab:02}) showed that our multimodal structure with all modalities included still outperformed same structure with single input modality; both video and audio positively contributed to the performance. We observed that the video-only system performed fairly well for both datasets, but the performance varied for the audio-only system. This discrepancy was caused by the difference in the audio quality. In the Olympic swimming dataset, the sound was clear and loud, while the trauma audio was often noisy and unclear. This was expected because the microphone array mounted on the side wall of the trauma room was far from the operation area (patient bed) and was close to an air vent. Noises from the hallway and air vent were sometimes louder than the medical equipment sounds and medical team's voices. In conclusion, using multiple input sources provided helpful information for process progress modeling, but only helped significantly if those inputs contained noticeable and representative features.

Finally, our system performance was compared to that of domain experts. To give the experts the same information, we started the video clips from the beginning of the video and stopped at a random time instance before the end. Using this approach, the domain experts could not use video's progress bar to estimate the overall completeness. The comparison was made using depth data from 10 cases. Experts outperformed our system in the progress estimation when given high resolution (1080p) RGB videos (Table \ref{tab:03}, Olympic swimming dataset). This outcome was expected, as the experts were the ones who generated the ground truth from RGB videos. Several other reasons contributed to the system's lower performance. The key reason is that our system did not use the state-of-the-art deep learning framework for image feature extraction; a deep ResNet might lead to better performance (we chose VGG-Net due to our limited computational power). In addition, the experts mentioned using extensive domain knowledge while making predictions. For example, in the trauma resuscitation dataset, the primary-survey phase should last less than 5 minutes. Training the system with both sensor data and domain knowledge could improve the system performance on process modeling in some domains. 

In addition, our system achieved performance comparable to experts for some tasks under constrained conditions (Table \ref{tab:03}, Resuscitation Dataset). For example, due to the privacy concerns, only depth cameras were allowed in the trauma room. Our results (Table \ref{tab:03}) showed that experts were more accurate even on depth videos, but also significantly slower (30 times slower than the machine) at predicting process phases. Domain experts used specific indicators for process phase detection, such as certain resuscitation activities or indicative tool usage. Recognizing specific activities requires both experience and domain knowledge. This knowledge may be hard for a person to learn only from depth video and audio. It was also difficult for the domain experts to estimate overall completeness, and almost impossible for them to estimate the remaining time while our system was able to estimate both.

\subsubsection{The Remaining Time Estimation}
\label{sec:0423}

\begin{figure}[!htb]
	\centering
	\includegraphics[width=0.7\textwidth]{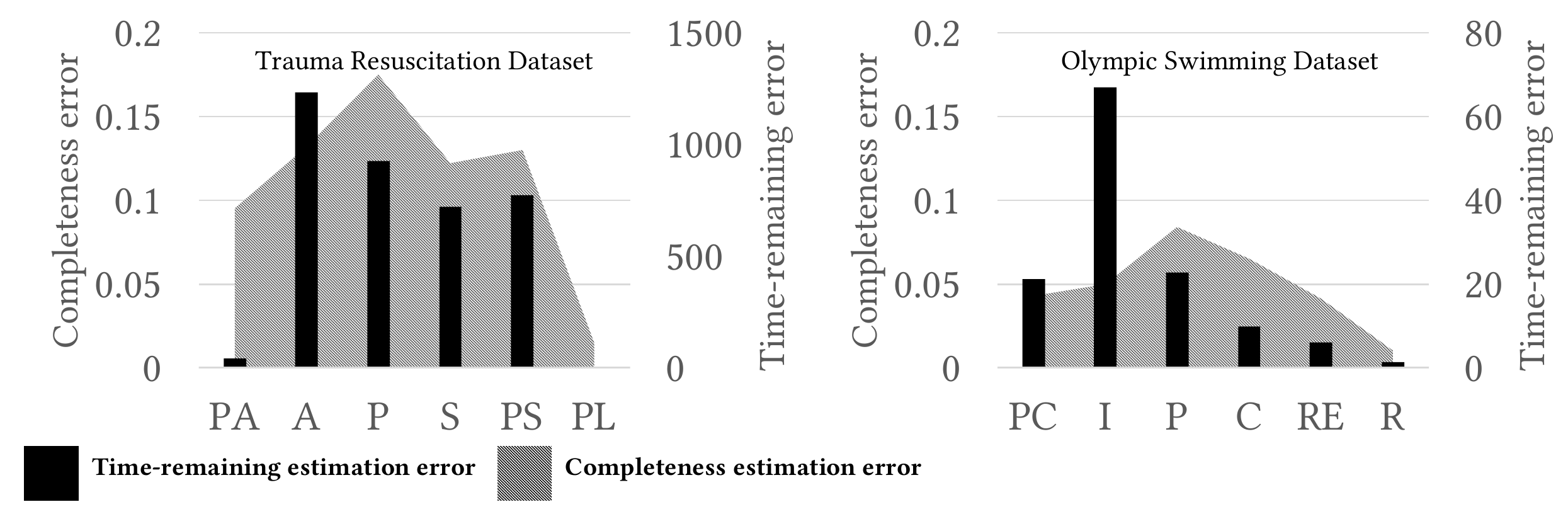}
	\captionsetup{justification=centering}
	\caption{Estimated time error for different datasets.}
	\label{fig:10}
\end{figure}

We also evaluated the average remaining-time estimation error, which was 7.5 minutes (14\% of total duration) for the trauma resuscitation dataset and 2.2 minutes (18\% of total duration) for the Olympic swimming dataset. Similar to the calculation of completeness error, we evaluated the remaining-time error by process phase (Fig. \ref{fig:10}). We found that the remaining-time estimation error was not associated with phase detection error, but with completeness estimation error. A small remaining-time estimation error in an earlier stage may lead to a large estimation error in the subsequent phases, as the estimated process speed will be multiplied by the percentage left in the process. Comparing the remaining-time estimation error of the trauma resuscitation dataset with the Olympic swimming dataset, we found that the error was also associated with process length (Fig. \ref{fig:06}, right); the longer processes had larger remaining-time estimation errors.

\begin{table}[!htb]
	\centering
	\setlength{\belowrulesep}{0pt}
	\setlength{\aboverulesep}{0pt}
	\caption{The Comparison of Process Phase Prediction by Domain Experts and Our System.}
	\label{tab:03}
	\begin{minipage}{\columnwidth}
		\begin{center}
			\begin{tabular}{|p{0.21\textwidth}|p{0.14\textwidth}|p{0.14\textwidth}|p{0.11\textwidth}|p{0.13\textwidth}|p{0.16\textwidth}|}
				\toprule
				\textit{\textbf{System}} & \textit{\textbf{Avg. Accuracy}} & \textit{\textbf{Avg. Precision}} & \textit{\textbf{Avg. Recall}} & \textit{\textbf{Avg. F1-Score}} & \textit{\textbf{Avg. time/frame}} \\ \midrule
				
				\textit{Resuscitation Expert} &            
				0.95 & 0.85 & 0.91 & 0.87 &
				$\approx$30s\\\midrule
				
				\textit{Resuscitation Machine}&         
				0.86 & 0.72 & 0.69 & 0.67&
				$<$1s\\\midrule
				
				\textit{Olympic Swim Expert} &            
				1 & 1 & 1 & 1 &
				$\approx$20s\\\midrule
				
				\textit{Olympic Swim Machine}&         
				0.88 & 0.69 & 0.66 & 0.68&
				$<$1s\\
				\bottomrule
			\end{tabular}
		\end{center}
	\end{minipage}
\end{table}

\subsection{Comparison of Phase Prediction to Previous Work}
\label{sec:043}

Since we were the first to attempt the estimation of completeness and remaining-time, we only compared the performance of process phase detection with our previous research \cite{li2016online}. The same datasets and training-testing splits were used for this comparison (6-phase trauma resuscitation \cite{li2016online}, 9-phase EndoTube dataset \cite{twinanda2017endonet} and 8-phase TUM LapChole dataset \cite{stauder2016tum}). Some evaluation scores were not reported in the original papers, e.g., the EndoNet handled first two phases as a single phase, and the averaging methods (average or weighted average) were not specified in the previous publications. Therefore, we reimplemented the models using the same training-testing splits \cite{twinanda2017endonet, stauder2016tum} to make the comparison fair. 

\begin{figure}[!b]
	\centering
	\includegraphics[width=0.8\textwidth]{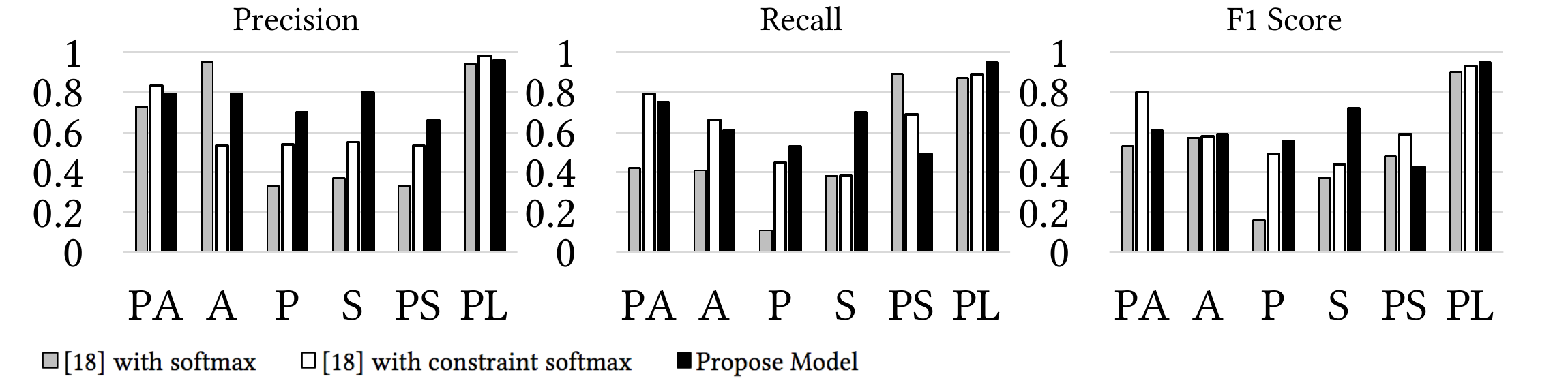}
	\captionsetup{justification=centering}
	\caption{Performance comparison of our previous (classification with constraint softmax \cite{li2016online}) and current models.}
	\label{fig:11}
\end{figure}

Our current model achieved significantly better performance compared to our previous model \cite{li2016online} (Fig. \ref{fig:11}).  A major drawback of the previous work was its reliance on spatial information and the omission of temporal associations \cite{li2016online}. Our previous model \cite{li2016online} could make incorrect predictions that contradicted common sense, e.g., predicting the primary-survey phase before patient arrival. To address this issue, we previously proposed a constraint softmax layer. Because the constraint softmax requires information across the window centered at the current time instance, the decision making must wait for the future data. With such a limitation, our previous system could only be used for post-event analysis. Our current system instead relies on a regression curve designed to generate ascending values between $[0,1]$, enforcing logically ordered predictions (i.e., patient arrival will happen before the primary-survey). Regression for completeness can be generated for every frame. Thus, our current system can work online. We further compared the phase prediction performance (Fig. \ref{fig:11}). The results showed that our current system outperformed the existing systems (Table \ref{tab:04}). Our current system maintained the similar F-score for pre-arrival and patient-leave phases, but significantly improved the scores for other phases. 

We next compared our current system using the EndoTube dataset \cite{twinanda2017endonet}, and the TUM LapChole dataset \cite{stauder2016tum}. Since both datasets contain repeating workflows (repeatedly performed surgical phases), our GMM based model did not work well on such datasets and could not compete with the previous research. This confirmed that the GMM cannot model nonlinear processes, because a single Gaussian distribution does not fit well to repeated phases. We slightly modified our current system by using another LSTM model instead of GMM for process phase detection and achieved slightly better performance compared to the existing systems (details in Section \ref{sec:05}).  

\begin{table}[!htb]
	\centering
	\setlength{\belowrulesep}{0pt}
	\setlength{\aboverulesep}{0pt}
	\caption{Phase prediction performance comparison using different input modalities.}
	\label{tab:04}
	\begin{minipage}{\textwidth}
		\begin{center}
			\begin{tabular}{@{}|p{0.287\textwidth}|p{0.042\textwidth}|p{0.19\textwidth}|p{0.151\textwidth}|p{0.032\textwidth}|p{0.04\textwidth}|p{0.04\textwidth}|p{0.047\textwidth}|@{}}
				\toprule
				\textbf{System}                                                                             & \textbf{Gen*}         & \textbf{Input Data}                     & \textbf{Model}                                       & \textbf{Acc.} & \textbf{Prec.} & \textbf{Rec.} & \textbf{F1-s.}  \\ \midrule
				Trauma Resuscitation Dataset&\multicolumn{7}{@{}l}{ }\\ \midrule
				Multimodal CNN {\cite{li2016online}}                                                        & Yes                   & Depth and audio                         & Multimodal CNN                                       & 0.62 & 0.60  & 0.51 & 0.50 \\ \midrule
				Multimodal CNN {\cite{li2016online}}                                                        & Yes                   & Depth and audio                         & Multimodal CNN $+$ constraint softmax                & 0.80 & 0.66  & 0.64 & 0.63 \\ \midrule
				\textit{Our model for linear processes}                                                                          & \textit{Yes}          & \textit{Depth and audio}                & \textit{CNN-LSTM Regression}                         & 0.86 & 0.72  & 0.69 & 0.67 \\ \midrule
				Endovis Dataset& \multicolumn{7}{@{}l}{ }\\ \midrule
				Endonet {\cite{twinanda2017endonet}}                                                        & Yes                   & Endoscope video                         & CNN+HHMM& 0.63 & 0.59  & 0.61 & 0.59  \\ \midrule
				\textit{Our model for linear processes}                                                                          & \textit{Yes}          & \textit{Endoscope video}                & \textit{CNN-LSTM Regression}                         & 0.66 & 0.55  & 0.61 & 0.51 \\ \midrule
				
				\textit{Our model for nonlinear process\^{}}                                                                          & \textit{Yes}          & \textit{Endoscope video}                & \textit{CNN-LSTM-LSTM}                         & 0.67 & 0.63  & 0.59& 0.61 \\ \midrule
				
				TUM LapChole Dataset&\multicolumn{7}{@{}l}{ }\\ \midrule
				CNN based approach {\cite{stauder2016tum}}                                                  & Yes                   & Laparoscopic video                      & AlexNet + time window                                & 0.72  & 0.61  & 0.66 & 0.63  \\ \midrule

				\textit{Our model for linear processes}                                                                          & \textit{Yes}          & \textit{Laparoscopic video}                & \textit{CNN-LSTM Regression}                         & 0.78 & 0.52  & 0.61 & 0.56 \\ \midrule
				
				\textit{Our model for nonlinear processes\^{}}                                                                        & \textit{Yes}          & \textit{Laparoscopic video}                & \textit{CNN-LSTM-LSTM}                         & 0.77 & 0.64  & 0.66 & 0.65 \\ \midrule
				
				Olympic Swimming Dataset &\multicolumn{7}{@{}l}{ }\\ \midrule
				\textit{Our model for linear processes}                                                                          & \textit{Yes}          & \textit{Video and audio}                & \textit{CNN-LSTM Regression}                         & 0.88 & 0.69  & 0.66 & 0.58 \\ \midrule
				{Other Previous Research}&\multicolumn{7}{@{}l}{ }                                                                                                                                                                                        \\ \midrule
				Phase detection from low-level activities {\cite{forestier2015automatic}}                   & No                    & Activity log                            & Decision Tree                                        & n/a  & 0.75  & 0.74 & 0.74 \\ \midrule
				Surgical phases detection {\cite{blum2008modeling}}                                         & No                    & Instrument signal                       & HMM                                                  & 0.83 & n/a   & n/a  & n/a  \\ \midrule
				Phase recognition using mobile sensors {\cite{bardram2011phase}}                            & No                    & Wearable sensor data                    & Decision Tree                                        & 0.77 & n/a   & n/a  & n/a  \\ \midrule
				Surgical workflow modeling {\cite{ padoy2012statistical}}                                              & No                    & Instrument signal                       & HMM                                                  & 0.84 & 0.85  & 0.84 & n/a  \\ \midrule
				Food preparation activities recognition {\cite{stein2013combining}}                                     & No                    & Video and accelerometer & Random Forest                                        & n/a  & 0.65  & 0.67 & n/a  \\
				\bottomrule
				\multicolumn{8}{@{}l}{*\footnotesize{Gen, If the system do not requires manually crafted feature as input.}}\\
				\multicolumn{8}{@{}l}{\footnotesize{\^{}Our model using LSTM structure instead of GMM for nonlinear process (Fig. \ref{fig:13}).}}
			\end{tabular}
		\end{center}
	\end{minipage}
\end{table}

We finally compared our system with existing phase-detection systems (Table \ref{tab:04}). We were unable to reproduce their results because they didn't publish their implementation or datasets, so we used their reported results in this paper for comparison. Since these existing systems were not evaluated on the same dataset, the reported evaluation scores might not reflect the system performance in each setting but could provide a general understanding of phase detection performance using different methods. We noticed that the system with the best performance \cite{forestier2015automatic} could predict high-level phases only using manually generated low-level activity logs. This is a significant limitation for real-world applications: automatically generated low-level activity logs using sensor data might contain errors, and may significantly influence the system performance. Moreover, our system is general enough to work with all types of medical resuscitations. Previous systems for process phase detection were trained to handle datasets with specific injury types \cite{padoy2012statistical,stauder2016tum,twinanda2017endonet}, but our current system is trained on cases with a variety of patient injury types, ages, and conditions. In terms of data collection, our system uses the economical, easily deployed, and commercially available depth camera with microphone array, while other early systems required human input, specific medical equipment, or obtrusive wearable sensors.

\begin{figure}[!b]
	\centering
	\includegraphics[width=0.8\textwidth]{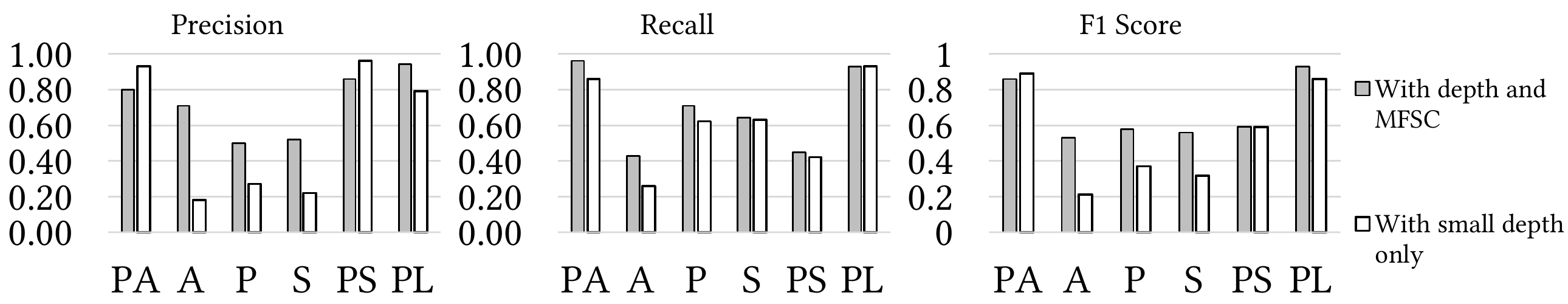}
	\captionsetup{justification=centering}
	\caption{Comparison of the phase prediction performance using small depth images as input and depth.}
	\label{fig:12}
\end{figure}

\section{Discussion}
\label{sec:05}

\subsection{Application --- Online Trauma Resuscitation Progress Detection}
\label{sec:051}

In addition to testing our system on recorded data, we deployed the system in a trauma room at CNMC. We mounted the entire system (a Kinect depth sensor and an Intel NUC mini PC) on a side wall of the trauma room (Fig. \ref{fig:06}, left). Since the mini PC uses a low-power CPU with no dedicated GPU, we had to shrink the input data size to $64\times64px$  to ensure that it is capable of running an online feedforward network. Considering that real-time MFSC feature extraction is computationally expensive, we used only depth video for resuscitation progress estimation. Using depth-only down-sampled video impacted the system performance (Fig. \ref{fig:12}). The performance, however, can be increased by replacing the mini PC by a more powerful computer or cloud computing solutions.

As mentioned earlier, we trained the model with \textit{Keras} using the \textit{TensorFlow} backend. Since a C$\#$ environment is required for online Kinect-based depth-video capturing, we established a local TCP-based communication to connect our C$\#$-based data capturing system to the feedforward network running on a Python-based terminal. We trained the model with depth frames captured during 50 resuscitation cases and tested the system with 10 resuscitation cases. The system achieved 8\% overall progress estimation error and 86\% phase detection accuracy (Fig. \ref{fig:12}). The accuracy is slightly lower than for the system that used greater-resolution depth frames with MFSC data as input, but the system performance could be improved by using greater-resolution video and replacing the mini PC by a computers with GPU. In addition, the deployment of our system for online estimation of trauma resuscitation progress revealed the challenges for real-world application, and the experience in solving them will help us better design the next generation of hardware and software.

\subsection{Limitations}
\label{sec:052}

Our current system works well with different datasets recorded in real-world scenarios. It still has some limitations. First, the completeness labels we used were only based on the time since the process started. As mentioned, the duration of a complex process can be affected by many factors. For example, during the post-secondary phase of trauma resuscitation, the availability of medical resources affects the waiting time, and a long waiting time does not necessarily indicate progress. Instead of labeling the data using time-based completeness, it might be reasonable to incorporate activity-performance indicators. For example, if the patient's left eye and right eye are both checked, then 5\% of the trauma resuscitation has been completed. Labeling the activities with the associated completeness is labor intensive. For complex medical processes, given different patient conditions, it is hard to quantify the contribution of an activity combination to the overall completeness. Further research in collaboration with the domain experts is necessary to design a time-activity-based phase recognition model.

Our current system assumes that the workflow process is linear. This system may not work well for nonlinear workflows, like the EndoTube and TUM LapChole datasets \cite{stauder2016tum,twinanda2017endonet}, because of our assumption that the GMM fits the distribution of a single occurrence for each phase. Nonlinear processes require representations with a mixture of normal distributions for each repeating phase. This problem can be addressed by modeling the process phase transition using transition topology or ``memory". Researchers previously used HMM or similar models for process phase detection\cite{padoy2012statistical}. However, HMM assumes that the current state is only associated with the directly preceding states, which may be inaccurate for process modeling: the current phase may have long-range dependencies with several previous phases. In addition, HMM does not fit well with our model structure because of its limitation in high-dimensional input observations, and HMM topologies for processes with many phases can be very complex. Unlike the GMM, the phase detection loss cannot be directly calculated by HMM and used for regression tuning.

\begin{figure}[!t]
	\centering
	\includegraphics[width=0.7\textwidth]{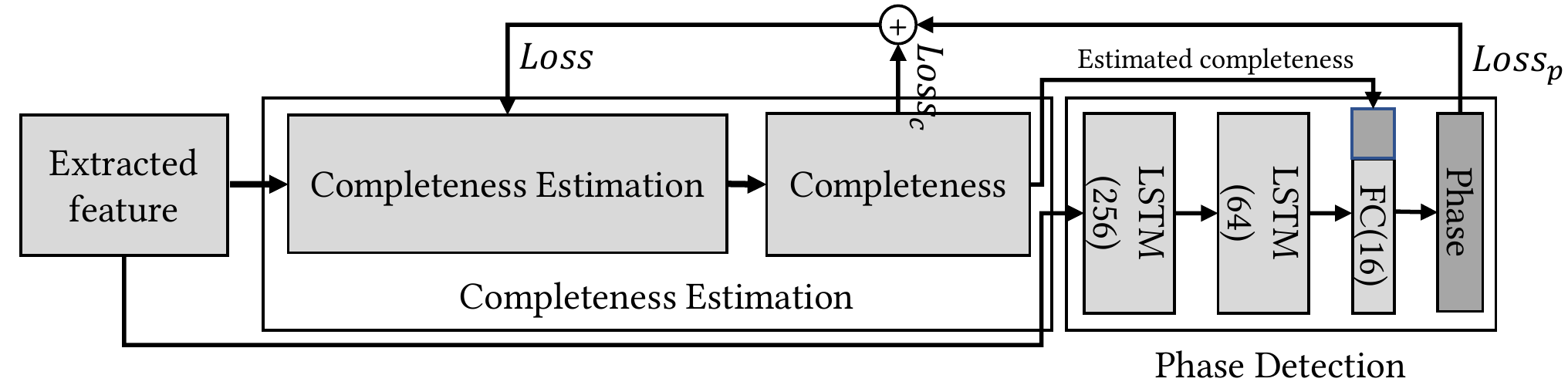}
	\captionsetup{justification=centering}
	\caption{Proposed CNN-LSTM-LSTM structure that could be used to estimate the progress of nonlinear processes.}
	\label{fig:13}
\end{figure}

An alternative to HMM is the LSTM, which has been shown to outperform HMM in sequential modeling like speech recognition \cite{graves2013speech}. In addition, the loss from an LSTM phase predictor could be used to tune the regressor. We propose an additional LSTM structure which takes both the extracted features and the estimated process completeness as the input for phase prediction (Fig. \ref{fig:13}).  Our preliminary results on the EndoVis dataset (with nine surgical phases) and TUM LapChole dataset (with eight surgical phases) showed that using LSTM for process phase detection achieved 4\% higher precision and 2\% higher F1-score compared to the GMM in our model (Table \ref{tab:04}). This performance improvement indicates that an LSTM structure could be used to model nonlinear processes. Our future work will include improving the CNN-LSTM-LSTM model for both linear and nonlinear process and testing it on additional datasets.

\subsection{Extension --- Smart Human Computer Interaction}
\label{sec:053}

Process progress estimation provides low-level information such as overall completeness and current process phase, which can be used to create smarter high-level human-computer interaction systems to improve work efficiency. Our system has many potential applications, for example, an online trauma resuscitation checklist. Currently, paper-based checklists are used in the trauma room to assist the trauma team with resuscitation management and to ensure that no required tasks are omitted. During fast-paced resuscitations the team members may forget to check off some items after the corresponding tasks are performed, they need to rely on their memory to determine whether these tasks were completed. Our system generates real-time progress estimation and phase predictions, which can be used contemporaneously to alert the trauma team to potentially omitted tasks. The resuscitation phase data (i.e. duration, start and end of each phase, and process execution time) can also be used to evaluate and score the resuscitation performance. As mentioned, our online resuscitation phase detection system is now working in a trauma room at the Children's National Medical Center. Our next step will be developing and testing a tablet-based digital checklist to be used during the trauma resuscitations. We will also study the benefits and drawbacks of the checklist by analyzing the statistics of resuscitation performance and receiving feedback from the trauma team.

\section{Conclusion}
\label{sec:06}

We introduced a system for estimating the progress of complex processes. We provided two real-world datasets collected from different commercially available sensors and used several published datasets to evaluate our system. Our system outperformed existing systems on two datasets from linear processes and achieved the performance comparable to existing systems on datasets from nonlinear processes. Our system can be applied in many real-world applications, providing essential information for advanced human-computer interaction. We deployed this system in a hospital trauma room for online resuscitation progress estimation and discussed potential extensions of the system. The paper contributes to the community:

\begin{enumerate}
	\item A regression-based neural network structure for process completeness estimation.
	\item The GMM-based approach and conditional loss that can be used with the regression model for classification tasks.
	\item The detailed system deployment instructions that can be used as a guideline to transfer the system to other fields.
	\item The trauma resuscitation and Olympic swimming datasets will be published with ground truth labels, and can be used for future studies.	
	\item A hybrid model that uses deep regression with LSTM instead of GMM to model nonlinear process with repeating phases. 
	
\end{enumerate}

\section{ACKNOWLEDGMENT}
	This research was supported by the National Institutes of Health under Award Number R01LM011834. The authors like to thank Colin Lea for providing the EndoTube dataset and the anonymous reviewers for their review and comments.

\bibliographystyle{acm}
\bibliography{sample-bibliography}

\end{document}